\documentclass{article}

\usepackage{arxiv}
\usepackage[utf8]{inputenc} % allow utf-8 input
\usepackage[T1]{fontenc}    % use 8-bit T1 fonts
\usepackage{hyperref}       % hyperlinks
\usepackage{url}            % simple URL typesetting
\usepackage{booktabs}       % professional-quality tables
\usepackage{amsfonts}       % blackboard math symbols
\usepackage{subfigure}
\usepackage{graphicx}
\usepackage{nicefrac}       % compact symbols for 1/2, etc.
\usepackage{microtype}      % microtypography
\usepackage{graphicx}
\usepackage{natbib}
\usepackage{doi}
\usepackage{multirow}
\usepackage{adjustbox}
\usepackage{booktabs}
\usepackage{amsmath}
\usepackage{bm}
\usepackage{comment}
\usepackage{tikz}
\usepackage{lipsum}

\title{Benchmarking of Query Strategies: Towards Future Deep Active Learning }

\author{ {Shiryu Ueno} \\
	Dept.\ of Natural Science and Technology\\
	Gifu University \\
	ueno@cv.info.gifu-u.ac.jp \\
 	\And
      {Yusei Yamada} \\
	Dept.\ of Natural Science and Technology\\
	Gifu University \\
	yyamada@cv.info.gifu-u.ac.jp \\
 	\And
       {Shunsuke Nakatsuka} \\
	Dept.\ of Natural Science and Technology\\
	Gifu University, Panasonic \\
	nakatsuka@cv.info.gifu-u.ac.jp \\
 	\And
        {Kunihito Kato} \\
	Dept.\ of Natural Science and Technology\\
	Gifu University \\
	kato.kunihito.k6@f.gifu-u.ac.jp \\
}

\renewcommand{\headeright}{Benchmark Paper}
\renewcommand{\undertitle}{Benchmark Paper}
\renewcommand{\shorttitle}{Benchmarking of Query Strategies: Towards Future Deep Active Learning}

\begin{document}
\maketitle

\begin{abstract}
In this study, we benchmark query strategies for deep actice learning~(DAL).
DAL reduces annotation costs by annotating only high-quality samples selected by query strategies. Existing research has two main problems, 
that the experimental settings are not standardized, making the evaluation of existing methods is difficult, and that most of experiments were conducted on the CIFAR or MNIST datasets. 
Therefore, we develop standardized experimental settings for DAL and investigate the effectiveness of various query strategies using six datasets, including those that contain medical and visual inspection images.
In addition, since most current DAL approaches are model-based, we perform verification experiments using fully-trained models for querying to investigate the effectiveness of these approaches for the six datasets.
Our code is available at \href{https://github.com/ia-gu/Benchmarking-of-Query-Strategies-Towards-Future-Deep-Active-Learning}.
\end{abstract}

\section{Introduction}
Deep learning requires a large annotated dataset.
The annotation cost in terms of time and money, is high when the target task requires professional experience, such as for medical and visual inspection images. 
Deep active learning~(DAL)~\cite{SurveyDeepActiveLearning,ComparativeSurveyDeepActiveLearning,DeepActiveLearningComputerVisionFuture} is a method that maintains or improves model performance while limiting the annotation cost.
DAL approaches can be divided into membership query synthesis~\cite{QueriesConceptLearning}, stream-based sampling~\cite{Dataefficientonlineclassificationsiamesenetworksactivelearning}, and pool-based sampling~\cite{EmployingEMPoolBasedActiveLearningTextClassification}.
In membership query synthesis, the model generates high-quality samples for learning.
In stream-based sampling, the model judges whether new samples in a data stream should be labeled. 
In pool-based sampling, the model selects high-quality samples for annotation from a pool of unlabeled data.
This study focuses on pool-based sampling, which is widely used in existing studies.

\begin{figure}[!tb]
    \begin{center}
    \includegraphics[keepaspectratio, width=80mm]{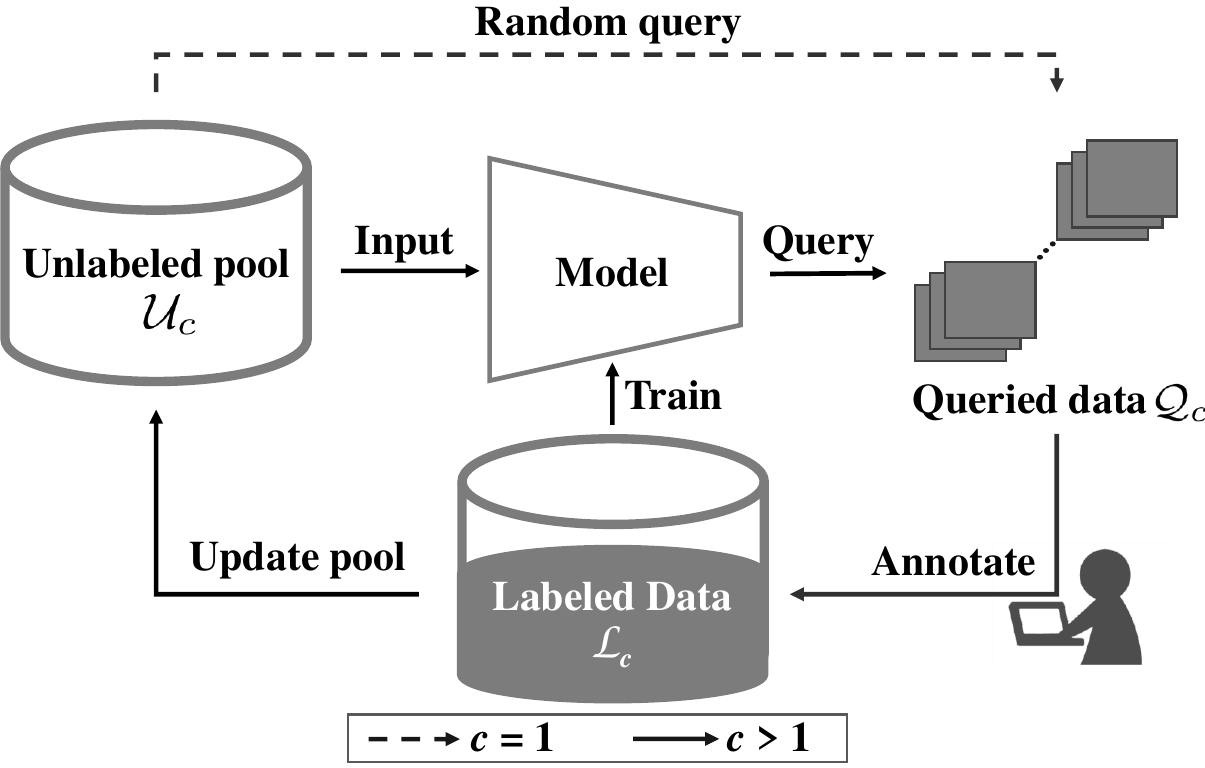}
    \caption{Framework of DAL. In cycle $c$, a model trained on labeled data $\mathcal{L}_c$ selects useful queried data $\mathcal{Q}_c$ from the unlabeled pool $\mathcal{U}c$. In cycle $c+1$, the process is repeated with $\mathcal{L}{c+1} = \mathcal{L}_c+\mathcal{Q}c$ and $\mathcal{U}{c+1} = \mathcal{U}_c - \mathcal{Q}_c$. In the first cycle~($c=0$), $\mathcal{Q}_c$ is randomly selected.}
    \label{pool-based-sampling}
    \end{center}
\end{figure}

Figure~\ref{pool-based-sampling} shows a typical DAL framework. 
As shown, DAL repeats the cycle of query, annotation, and training under the annotation budget or target accuracy. Querying refers to the selection of samples.
By constructing training data with high-quality samples in each cycle, DAL can quickly reach the target accuracy, reducing the annotation cost.

In studies on DAL, the hyperparameters related to the number of training data~(e.g., number of queried data $|\mathcal{Q}_c|$ or total number of cycles $C$) and other experimental settings~(e.g., model initialization settings at the beginning of each cycle) are not standardized.
In addition, the random seed is not fixed in some experiments, making a performance comparison of existing methods difficult.
%. 
Many of the reported experiments were performed on the CIFAR~\cite{LearningMultipleLayersFeaturesTinyImages} or MNIST~\cite{Gradientbasedlearningapplieddocumentrecognition} dataset, both of which are homogeneous~(i.e., balanced and clean). 
However, DAL is expected to be used with more practical datasets~(e.g., medical or visual inspection images), which are imbalanced or noisy.
Therefore, in this study, we develop a standardized experimental settings for DAL and investigate the effectiveness of various query strategies with six datasets, including those that contain medical or visual inspection images.

In addition, as shown in Figure ~\ref{pool-based-sampling}, $\mathcal{Q}_c$ is queried by models trained on labeled data $\mathcal{L}_c$.
However, when $|\mathcal{L}_c|$ is small, the model may not be sufficiently trained and thus may not extract good features.
In this case, query strategies based on features do not work well, and high-quality $\mathcal{Q}_c$ cannot be obtained.
To provide guidelines for future DAL studies, we perform verification experiments to compare the results of the currently used framework~(Figure ~\ref{pool-based-sampling})~with the results of a framework that uses a fully-trained model for queries.
A fully-trained model can query $\mathcal{Q}_c$ independent of the value of $|\mathcal{L}_c|$.
Through the verification experiments, we investigate the effectiveness of the current DAL approach, which depends on the number of training data.

Our contributions are as follows.

\begin{itemize}
    \item We show the inconsistency of experimental settings in reported experiments on DAL and the difficulty of comparing the performance of existing methods due to this inconsistency.
    To overcome this problem, we develop standardized experimental settings for DAL and use them to benchmark existing methods.
    \item In addition to homogeneous datasets, we benchmark existing methods with datasets of medical or visual inspection images. We also perform experiments using pre-training with self-supervised learning to investigate the effectiveness of self-supervised learning for DAL.
    \item We perform verification experiments with a fully-trained model for querying $\mathcal{Q}_c$ and compare the results with those in the benchmarking experiments. 
    We show that developing a query strategy based on the current DAL approach and verifying its effectiveness for a homogeneous dataset is difficult and that doing so far a non-homogeneous dataset would be beneficial. We also show the importance of considering the characteristics of the target dataset when developing a query strategy.
\end{itemize}

    % As a result, we show that proposing a new query strategy based on the current DAL approach and show its effectiveness on a homogeneous dataset is difficult, but to propose a query strategy that takes into account the characteristics of a non-homogeneously organized dataset with properties such as imbalanced. 

\section{Related Work}

\subsection{Query Strategies}
DAL aims to reduce annotation costs by querying only high-quality samples with query strategies. The query strategies can be categorized into uncertainty-based, representative/diversity-based, and hybrid strategies.
 
\subsubsection{Uncertainty-based query strategies}
Uncertainty-based query strategies select uncertain samples by using sample complexity or ambiguity. This is based on the concept that uncertain samples contain useful information for the model and should be labeled.
Typical methods use the predicted probability of the model~\cite{HeterogeneousUncertaintySamplingSupervisedLearning,IdentifyingWronglyPredictedSamplesMethodActiveLearning}, the predicted probability distribution of Bayesian neural network~\cite{Bayesianneuralnetworksintroductionsurvey,BALanCeDeepBayesianActiveLearningEquivalenceClassAnnealing,DeepBayesianActiveLearningImageData}, or the consistency of the model prediction~\cite{ConsistencybasedActiveLearningObjectDetection,CollaborativeIntelligenceOrchestrationInconsistencyBasedFusionSemiSupervisedLearningActiveLearning}.
One major problem with uncertainty-based query strategies is that they tend to query samples near the boundary, and thus they select similar samples.

~\subsubsection{Representative/diversity-based query strategies}
Representative/diversity-based query strategies select samples to broaden the distribution of the query data. This avoids the selection of similar samples, which is a problem with uncertainty-based query strategies.
Typical methods use clustering~\cite{Activelearningusingpreclustering,DeepActiveLearningLongTail}, adversarial sampling~\cite{AdversarialActiveLearningDeepNetworksMarginBasedApproach,GenerativeAdversarialActiveLearning}, or generative models~\cite{VariationalAdversarialActiveLearning}.
One major problem with representative/diversity-based methods is that they do not consider the uncertainty of data, so the model sometimes queries easy samples. In addition, methods that use clustering or adversarial sampling are computationally expensive compared with uncertainty-based methods.

\subsubsection{Hybrid-based query strategies}
Hybrid-based query strategies combine uncertainty-based methods and representative/diversity-based methods~\cite{DeepActiveLearningUnifiedPrincipledMethodQueryTraining,DeepSimilarityBasedBatchModeActiveLearningExplorationExploitation}. This approach cancels out the disadvantages of each type of method.
Some hybrid-based methods combine existing uncertainty-based methods and representative/diversity-based methods~\cite{DiverseminibatchActiveLearning}. 
As with representative/diversity-based methods, these methods often require clustering or adversarial sampling and are computationally expensive.

\subsubsection{Issues with existing work}
In previous studies, hyperparameters related to the number of training data and other experimental settings are not standardized. This leads to different performance even if $|\mathcal{L}_c|$ is the same~\cite{MakingYourFirstChoiceAddressColdStartProblemVisionActiveLearning,ColdstartActiveLearningRobustOrdinalMatrixFactorization,AddressingItemColdstartProblemAttributedrivenActiveLearning}.
In addition, the random seed was not fixed in some experiments, and thus comparing the performance of existing methods is difficult. To overcome these problems, we develop standardized experimental settings for DAL and use them to benchmark existing methods.

\subsection{DAL for Medical or Visual Inspection Images}
Most previous studies proposed query strategies and verified them using the CIFAR or MNIST dataset. Query strategies for medical or visual inspection images have also been proposed~\cite{Patientawareactivelearningfinegrainedoctclassification,InformationGainSamplingActiveLearningMedicalImageClassification}.
For example, Logan et al.~\cite{DECALDEployableClinicalActiveLearning} proposed a method for querying a certain number of samples for each patient from their optical coherence tomography~(OCT)~\cite{OCTIDOpticalCoherenceTomographyImageDatabase} and X-Ray~\cite{IdentifyingMedicalDiagnosesTreatableDiseasesImageBasedDeepLearning}.
Pimentel et al.~\cite{DeepActiveLearningAnomalyDetection} proposed unsupervised to active inference~(UAI), which defines an anomaly metric for each sample using embedding expressions and anomaly scores with a denoising auto encoder~\cite{Extractingcomposingrobustfeaturesdenoisingautoencoders}. They proposed a method for querying samples based on UAI.

The motivation for these studies was the assumption that existing methods proposed for the CIFAR and other homogeneous datasets do not work for medical or visual inspection images. 
However, to the best of our knowledge, this assumption has not been verified. Therefore, in this paper, we investigate the effectiveness of query strategies for medical or visual inspection images.

\subsection{DAL for Imbalanced, Out-of-Distribution, or Noisy Data}
When considering the practical use of DAL, the target datasets may include imbalanced~\cite{Imbalancedlearnpythontoolboxtacklecurseimbalanceddatasetsmachinelearning}, out-of-distribution(OoD)~\cite{Canyoutrustyourmodeluncertaintyevaluatingpredictiveuncertaintydatasetshift}, or noisy labels~\cite{Learningnoisylabels}.
Therefore, some methods for datasets containing such labels have been proposed~\cite{ClassBalancedActiveLearningImageClassification,ForgetfulActiveLearningSwitchEventsEfficientSamplingOutofDistributionData,DeepActiveLearningNoiseStability}.
For example, Killamsetty et al.~\cite{GLISTERGeneralizationbasedDataSubsetSelectionEfficientRobustLearning} proposed a method for querying a subset of data to maximize the log-likelihood for validation set using a submodular function~\cite{Submodularfunctionsoptimization} and showed its performance for imbalanced or noisy data.
Kothawade et al.~\cite{SimilarSubmodularinformationmeasuresbasedactivelearningrealisticscenarios} proposed the transformation of submodular mutual information~(SMI)~\cite{onlinesubmodularcoverproblem,Submodularcombinatorialinformationmeasuresapplicationsmachinelearning} into a form corresponding to imbalanced, OoD, or redundant data, and queried samples to maximize the SMI.

Because many previous studies verified their methods using artificial non-homogeneous datasets based on the CIFAR or MNIST dataset, which are homogeneous, it is unclear whether they work on datasets that contain actual non-homogeneous data, such as medical or visual inspection images.
Furthermore, most existing methods require prior knowledge of the imbalanced or OoD characteristics in the datasets.
However, in the DAL framework, the datasets are not initially annotated, making it impossible to know in advance their characteristics. Consequently, these methods are not practical in this context.

\subsection{DAL with Semi-Supervised Learning}
Semi-supervised learning(semi-SL)~\cite{Semisupervisedlearning,surveysemisupervisedlearning} uses both labeled and unlabeled data. 
This matches the DAL framework and thus methods that combine DAL and semi-SL have been proposed.
For example, Gao et al.~\cite{ConsistencybasedSemisupervisedActiveLearningMinimizingLabelingCost} used the MixMatch~\cite{Mixmatchholisticapproachsemisupervisedlearning} framework of semi-SL to prepare a large amount of augmented data and query samples based on the consistency of model prediction.
Yuan et al.~\cite{Activematchendtoendsemisupervisedactiverepresentationlearning}, proposed a combination of the FixMatch~\cite{FixmatchSimplifyingsemisupervisedlearningconsistencyconfidence} framework and contrastive learning~\cite{simpleframeworkcontrastivelearningvisualrepresentations,Momentumcontrastunsupervisedvisualrepresentationlearning} to train models using both labeled and unlabeled data. Samples are queried using the trained models.
MixMatch and FixMatch require strong data augmentation and are thus ineffective for non-natural images. Due to this limitation, we do not utilize a combination of semi-SL and DAL in this study.

\subsection{DAL with Self-Supervised Learning}
Self-supervised learning~(self-SL)~\cite{surveycontrastiveselfsupervisedlearning} uses unlabeled data with mechanically generated labels, called Pretext Task~\cite{SelfieSelfsupervisedPretrainingImageEmbedding,Selfsupervisedlearningpretextinvariantrepresentations} or contrastive learning.
Since self-SL does not require labeled data for learning, it is often used as a pre-training method. Furthermore, methods that combine self-SL with DAL have been proposed.
For example, Yi et al.~\cite{PT4ALUsingSelfSupervisedPretextTasksActiveLearning}, proposed a method that utilizes angle prediction as a pretext task for pre-training, and queries samples based on the loss of the pretext task.
Caramalau et al.~\cite{MoBYv2ALSelfsupervisedActiveLearningImageClassification}, replaced the encoder of MoBY~\cite{SelfSupervisedLearningSwinTransformers}, a self-SL method proposed for the Swin-Transformer~\cite{SwintransformerHierarchicalvisiontransformerusingshiftedwindows}, with a CNN. They then modified the framework to enable MoBY to learn from not only unlabeled data but also labeled data. Samples are queried using the trained model.

However, these studies performed experiments using the CIFAR or MNIST dataset. In this paper, we investigate the effectiveness of self-SL with DAL not only for homogeneous datasets, but also for non-homogeneous datasets by performing experiments using pre-training with SimSiam~\cite{ExploringSimpleSiameseRepresentationLearning}.

\section{Experiments: Benchmarking of Query Strategies}
\subsection{Datasets}
In this paper, we use CIFAR10 for natural images, EuroSAT for satellite images~\cite{EuroSATNovelDatasetDeepLearningBenchmarkLandUseLandCoverClassification},
OCT and BrainTumor for medical images~\cite{braintumordatasetfigshareDataset},
GAPs and KolektorSDD2 for visual inspection images~\cite{ImprovingVisualRoadConditionAssessmentExtensiveExperimentsExtendedGAPsDataset,Mixedsupervisionsurfacedefectdetectionweaklyfullysupervisedlearning}.

CIFAR10 is a natural image dataset with 50,000 training data and 10,000 test data, and consists of 10 classes with an equal number of data per class. The image size is $32\times32$ pixels. We use RandomCrop and RandomHorizontalFlip with a probability of $p=0.5$ as data augmentation.

\begin{table}[!tb]
\caption {\textmd{Distribution of training and test data for EuroSAT dataset}}
\vspace{3.0mm}
\label {eurosat_distribution}
\centering
\scalebox{1}{
    \begin{tabular}{lcc}
    \hline
Classes                      & training data    & test data      \\ \hline
Industrial Buildings         & 2,037            & 463            \\
Residential Buildings        & 2,444            & 556            \\
Annual Crop                  & 2,445            & 555            \\
Permanent Crop               & 2,037            & 463            \\ 
River                        & 2,037            & 463            \\ 
Sea \& Lake                   & 2,444            & 556            \\ 
Herbaceous Vegetation        & 2,445            & 555            \\ 
Highway                      & 2,037            & 463            \\ 
Pasture                      & 1,629            & 371            \\ 
Foreset                      & 2,445            & 555            \\ \hline
Summary                      & 22,000           & 2,200          \\ \hline
    \end{tabular}
}
\end{table}

EuroSAT is a satellite image dataset consisting of 10 classes and a total of 27,000 data, with a relatively equal number of data per class. The image size is $64 \times 64$ pixels. In accordance with previous studies~\cite{RemoteSensingImageSceneClassificationMeetsDeepLearningChallengesMethodsBenchmarksOpportunities,Learningtransferablevisualmodelsnaturallanguagesupervision}, we use 22,000 data for training and 5,000 data for testing. The distribution of the number of data per class after splitting EuroSAT is shown in Table~\ref{eurosat_distribution}. We use RandomVerticalFlip with a probability of $p=0.5$ as data augmentation.

OCT is a medical image dataset of gray-scale cross-sectional images of retinas with 83,484 training data and 968 test data, and consists of four classes. OCT is an imbalanced dataset. The image size is not uniform, so we resized the images to $224\times224$ pixels.
\begin{figure}[!t]
\centering
    \subfigure[~(a)~Results for full training data]
    {\includegraphics[width=0.45\linewidth]{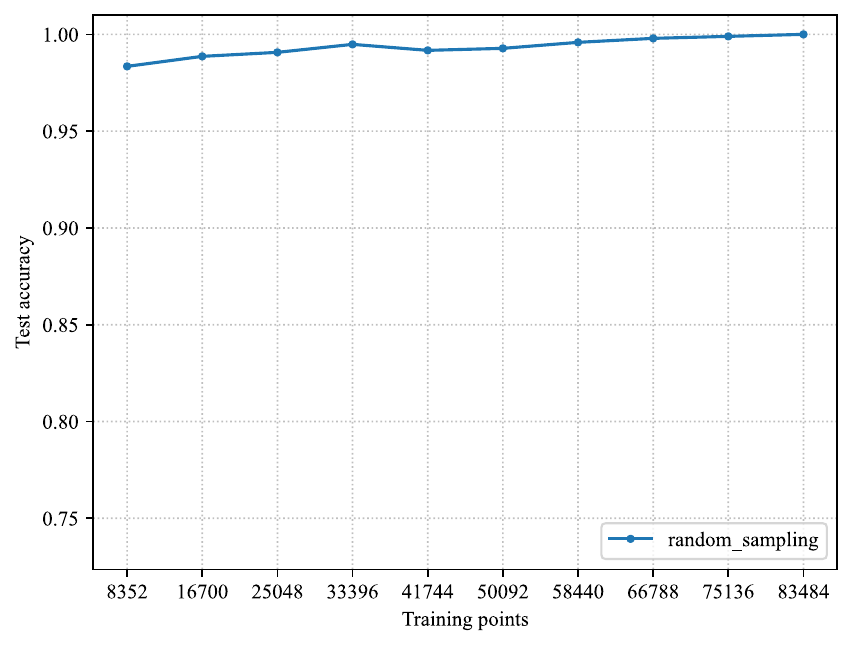}}
	\subfigure[~(b)~Results for reduced training data]
    {\includegraphics[width=0.45\linewidth]{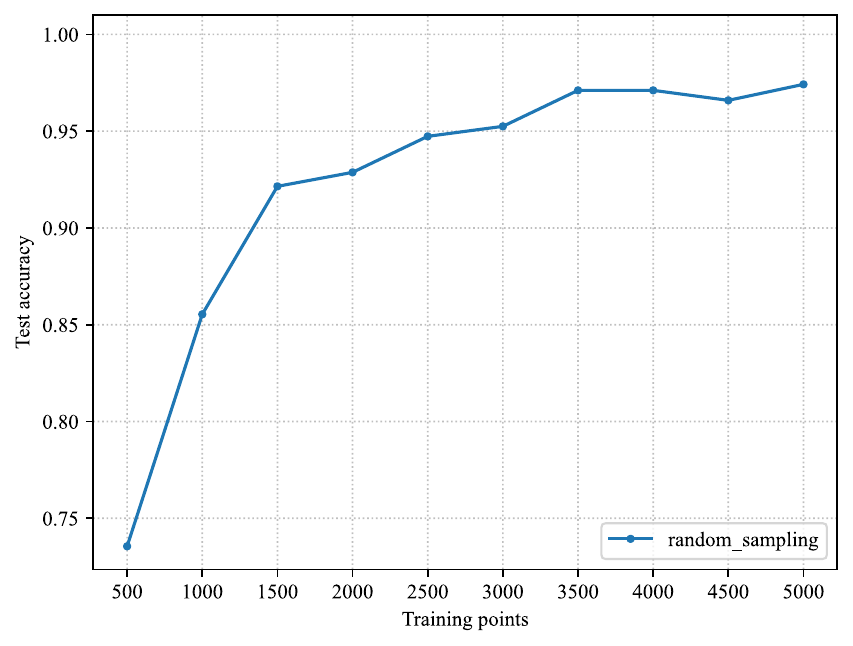}}
    \caption{Results obtained with various numbers of training data for OCT}
    \label{oct_pre_result}
\end{figure}
Figure~\ref{oct_pre_result}~(a) shows the results of DAL obtained using the full training data of OCT. 
As shown, when using the full training data, the accuracy in the first cycle is approximately 98\%; it stays within 98-100\% in subsequent cycles. 
Therefore, the difference in accuracy between the methods is at most about 1\%. This difference can be considered to be error, making it difficult to compare the methods. 
To address this issue, we randomly reduced the training data from 83,484 to 5,000 data while maintaining the distribution of the classes.
Figure~\ref{oct_pre_result}~(b) shows the results of DAL obtained with the reduced dataset.
As shown, the improvement in accuracy is significant. It is likely that the difference in accuracy between the methods is greater than the error. The distribution of the number of training data per class after the reduction of OCT is shown in Table~\ref{oct_distribution}. 

\begin{table}[!tb]
\caption {\textmd{Distribution of full and reduced training data for OCT dataset}}
\vspace{3.0mm}
\label {oct_distribution}
\centering
\scalebox{1}{
    \begin{tabular}{lcc}
    \hline
    Classes         & full data         & reduced data      \\ \hline
    NORMAL          & 26,315            & 1,576             \\
    CNV             & 37,205            & 2,229             \\
    DME             & 11,348            & 679               \\
    DRUSEN          & 8,616             & 516               \\ \hline
    Summary         & 83,484            & 5,000             \\ \hline
    \end{tabular}
}
\end{table}
\begin{table}[!tb]
\caption {\textmd{Distribution of training and test data for BrainTumor dataset}}
\vspace{3.0mm}
\label {braintumor_distribution}
\centering
\scalebox{1}{
    \begin{tabular}{lcc}
    \hline
    Classes         & training data         & test data      \\ \hline
    Glioma          & 1,139                 & 287            \\
    Meningioma      & 542                   & 166            \\
    Pituitary Tumor & 740                   & 190            \\ \hline
    Summary         & 2,421                 & 643            \\ \hline
    \end{tabular}
}
\end{table}

BrainTumor is a medical image dataset of gray-scale magnetic resonance imaging images of the brain with a total of 3,064 data, and consists of three classes. BrainTumor is an imbalanced dataset. The image size is $512\times512$ pixels, which is rather large, so we resized the images to $256\times256$ pixels. BrainTumor is provided with five-fold cross-validation. Therefore, in this paper, we split the data into training and test data with a ratio of 4:1.
The distribution of the number of training data per class after splitting BrainTumor is shown in Table~\ref{braintumor_distribution}.

GAPs is a dataset of visual inspection images of asphalt surfaces taken in Germany. There are three types of data: GAPs 10m, GAPs v2, and GAPs v1. In this paper, we use GAPs v2, which is available for image classification.
GAPs consists of 2,468 gray-scale images with a size of $1920\times1080$ pixels. The images can be divided into patches of arbitrary sizes. In accordance with previous studies~\cite{ImprovingVisualRoadConditionAssessmentExtensiveExperimentsExtendedGAPsDataset}, the patch size was set to $64\times64$ pixels.
GAPs includes 50,000 training data and 10,000 test data. It is strongly imbalanced. Moreover, GAPs includes noise-labeled data.

KolektorSDD2 is a dataset of gray-scale visual inspection images of the surfaces of manufactured products. It has 2,331 training data and 1,004 test data. KolektorSDD2 is strongly imbalanced. The image size is $230\times630$ pixels, so we resized the images to $224\times224$ pixels.
KolektorSDD2 is a dataset for segmentation, so only the training data annotations for classification are provided.
Therefore, we annotated the 110 NG~(anomaly) samples in the test data based on~\cite{Mixedsupervisionsurfacedefectdetectionweaklyfullysupervisedlearning} and segmentation ground truth.

Using these datasets, we investigate the effectiveness of existing query strategies using medical and visual inspection images. 
Furthermore, since BrainTumor, OCT, GAPs, and KolektorSDD2 are imbalanced datasets, we investigate the effectiveness of these strategies on imbalanced data. 
Moreover, since GAPs contains noise-labeled data, we investigate the effectiveness of existing query strategies on a noisy dataset using results from GAPs.

\subsection{Methods}
In this paper, we use Random Sampling, Entropy Sampling, BatchBALD~\cite{BatchBALDEfficientDiverseBatchAcquisitionDeepBayesianActiveLearning}, k-means Sampling, Core-set~\cite{ActiveLearningConvolutionalNeuralNetworksCoreSetApproach}, BADGE~\cite{DeepBatchActiveLearningDiverseUncertainGradientLowerBounds}, and Cluster Margin~\cite{BatchActiveLearningScale}.

Random Sampling randomly queries samples from the unlabeled pool $\mathcal{U}_c$. As other methods query samples strategically, Random Sampling is considered to be a baseline for evaluating the effectiveness of query strategies.

Entropy Sampling queries samples with large values for Equation~(\ref{eq:entropy sampling}).
\begin{align}
H = p(\bm{y}|\bm{x})\text{log}p(\bm{y}|\bm{x})
\label{eq:entropy sampling}
\end{align}
where $\bm{x}$ is an unlabeled data, $\bm{y}$ is the class label, and $p(\bm{y}|\bm{x})$ is the model's predicted probability.

BatchBALD queries samples that maximize the mutual information of the predicted probability distribution of a Bayesian neural network, as shown in Equation~(\ref{eq:batchbald}). 
\begin{align}
H(\bm{y}|\bm{x}, \mathcal{L}_{c-1}) - E_{p(\omega|\mathcal{L}_{c-1})}[H(\bm{y}|\bm{x}, \omega, \mathcal{L}_{c-1})]
\label{eq:batchbald}
\end{align}
where $E$ is the expected value and $\omega$ is the model's parameter distribution. In this study, we use mote-carlo~(MC) dropout~\cite{DropoutBayesianApproximationRepresentingModelUncertaintyDeepLearning} instead of a Bayesian neural network.
While normal dropout is only performed during training, MC dropout is also performed during testing. By repeatedly performing testing with dropout and averaging the results, it is possible to approximate the distribution of each weight and estimate the distribution of the test results.
The number of MC dropout iterations is a hyperparameter, but since it is impossible to perform hyperparameter optimization~(i.e., parameter tuning) in DAL, we fix the number of MC dropout iterations to 40, following a previous study~\cite{EffectiveEvaluationDeepActiveLearningImageClassificationTasks}.

k-means Sampling performs k-means clustering~\cite{METHODSCLASSIFICATIONANALYSISMULTIVARIATEOBSERVATIONS}, with the number of clusters set to the number of query data $|\mathcal{Q}_c|$, and queries the sample closest to the centroid of each cluster.

Core-set uses the k-center problem~\cite{FacilitylocationConceptsmodelsalgorithmscasestudies} to query the subset of $\mathcal{U}_{c-1}$. However, solving the k-center problem is NP-hard, so Core-set uses the k-center greedy approach to query samples that satisfy Equation~(\ref{eq:coreset}).
\begin{align}
\text{argmax}_{i\in \mathcal{U}_{c-1}}\text{min}_{j\in \mathcal{L}_{c-1}}d(f_i, f_j)
\label{eq:coreset}
\end{align}
where $d$ is a distance function and $f$ represents features.

BADGE performs k-means++ Clustering~\cite{kmeansadvantagescarefulseeding} using the gradient vectors of unlabeled data. 
k-means++ Clustering alleviates the problem of initialization dependence in k-means Clustering by performing initialization using distances between data.
By using gradient vectors, BADGE can consider sample uncertainty, and by using k-means++ Clustering, it can also consider sample diversity.

\begin{figure}[!tb]
    \begin{center}
    \includegraphics[keepaspectratio, width=85mm]{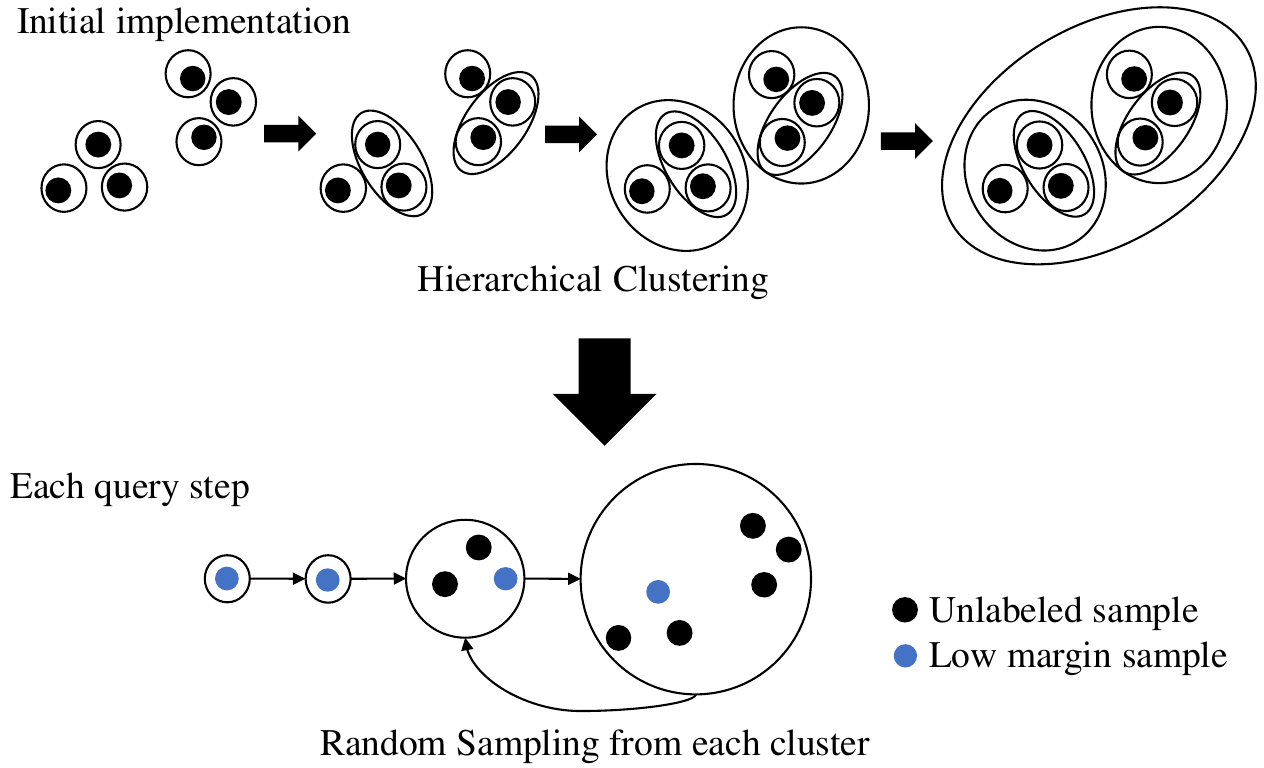}
    \caption{Framework of Cluster Margin}
    \label{cluster_margin}
    \end{center}
\end{figure}

The framework of Cluster Margin is shown in Figure~\ref{cluster_margin}. First, hierarchical clustering is performed once. 
Then, during each query, Random Sampling is performed from small clusters in the order of small values for Equation~\ref{eq:margin}.
\begin{align}
\text{max}_{i\in \mathcal{U}_{c-1}}(p(\bm{y}|\bm{x}_i))-\text{max}_{j\ne i, j\in \mathcal{U}_{c-1}}(p(\bm{y}|\bm{x}_j))
\label{eq:margin}
\end{align}
By using hierarchical clustering, Cluster Margin can consider sample diversity, and by using Equation~\ref{eq:margin}, it can also consider sample uncertainty. In addition, since hierarchical clustering is performed only once in Cluster Margin, the computational cost is relatively low.

We use the seven methods described above to investigate the effectiveness of uncertainty-based, representative/diversity-based, and hybrid query strategies for each dataset.
In addition, there are query strategies that use a module to predict the loss~\cite{LearningLossActiveLearning} or a generative model~\cite{Taskawarevariationaladversarialactivelearning}. 
However, since these methods have different model structures than those of other query strategies, it is difficult to fairly evaluate their performance. Therefore, they are not used in this study.

To investigate the effectiveness of self-SL for DAL, we conducted experiments using SimSiam for pre-training.
SimSiam is a type of contrastive learning, where two different data augmentations are applied to each data, and learning is performed using twice as many augmented data.
Augmented data from the same original data are positive pairs and those from different original data are negative pairs. 
Contrastive learning learns to increase the similarity between positive pairs and decrease the similarity between negative pairs. 
Many contrastive learning methods use a large batch size for similarity calculation with negative pairs. However, SimSiam learns only from positive pairs, making stable learning possible even with a small batch size.
\subsection{Experimental Settings}
\subsubsection{Settings for DAL}
For each dataset, $|\mathcal{Q}_c|$ is 10\% of the entire training data and $C$ is 10.
For example, for CIFAR10, the model randomly queries 5,000 images in the first cycle, then strategically queries another 5,000 images and learns with a total of 10,000 labeled data in the next cycle.
However, since BrainTumor and KolektorSDD2 have fractions in their training data, we randomly query the fractions in the first cycle.
At the beginning of each cycle, we initialize the model weights. When we use pre-training, we initialize the model weights with pre-trained weights. We refer to these settings as scratch and self-SL, respectively.
For each dataset, we evaluate the results using the average for five seeds.
The evaluation metric is accuracy for the test data, except for GAPs and KolektorSDD2, where accuracy evaluation is difficult due to a strong imbalance in visual inspection images.
Therefore, for these two datasets, we use F1-score~\cite{AccuracyFScoreROCFamilyDiscriminantMeasuresPerformanceEvaluation} for the test data.
GAPs training is a six-class classification~(one normal class and five anomaly classes). For testing, we treat the five anomaly classes as one class, resulting in a two-class classification.

\subsubsection{Settings for deep learning}
For all datasets, we use ResNet18~\cite{DeepResidualLearningImageRecognition} as the base model. For CIFAR10, for which the image size is very small~($32 \times 32$ pixels), we modify the size of the convolutional kernel in the first layer from $7 \times 7$ to $3 \times 3$. 
We set the batch size to 20, use the stochastic gradient descent optimizer with a learning rate of 0.01, a momentum of 0.9, and a weight decay of 0.0005, and train the model for 200 epochs. We also apply cosine decay to the learning rate.

\subsubsection{Settings for SimSiam}
For SimSiam, we use the output of ResNet18 up to just before the final classification layer as the encoder. For CIFAR10, we modify the convolutional kernel in the first layer from $7\times7$ to $3\times3$.
We set the batch size to 128, use the stochastic gradient descent optimizer with a learning rate of 0.05, a momentum of 0.9, and a weight decay of 0.0005, and train the model for 800 epochs. We also apply cosine decay to the learning rate.
The data augmentations used for SimSiam are ResizedCrop, ColorJitter, Grayscale, GaussianBlur, and HorizontalFlip, similar to BYOL\cite{Bootstrapyourownlatentanewapproachselfsupervisedlearning}. The probabilities are $q=0.8$ for ColorJitter, $q=0.2$ for Grayscale, and $q=0.5$ for the other augmentations. We do not use Grayscale for OCT, BrainTumor, GAPs, and KolektorSDD2, for which the original images are gray-scale.
\subsection{Benchmark Results}\label{result of benchmark}
\begin{figure*}[!tb]
\centering
    \subfigure[CIFAR10 scratch]{\includegraphics[width=0.245\linewidth]
    {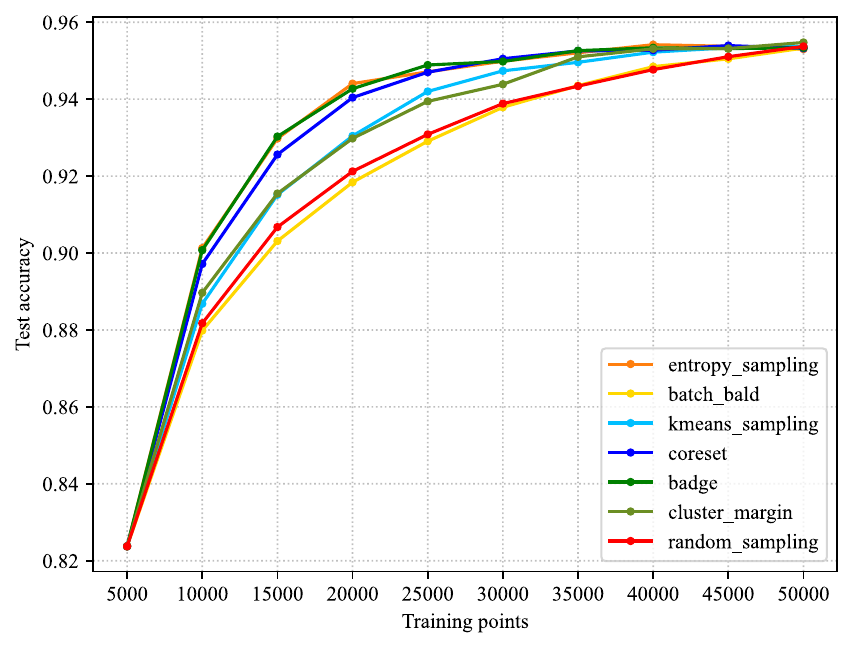}}
	\subfigure[CIFAR10 Self-SL]{\includegraphics[width=0.245\linewidth]{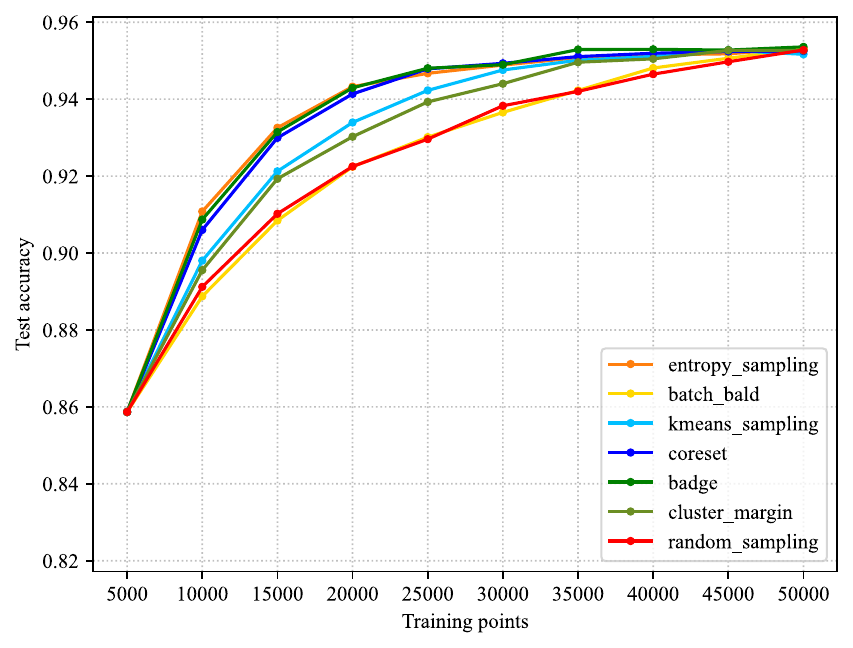}}
     \subfigure[EuroSAT scratch]{\includegraphics[width=0.245\linewidth]{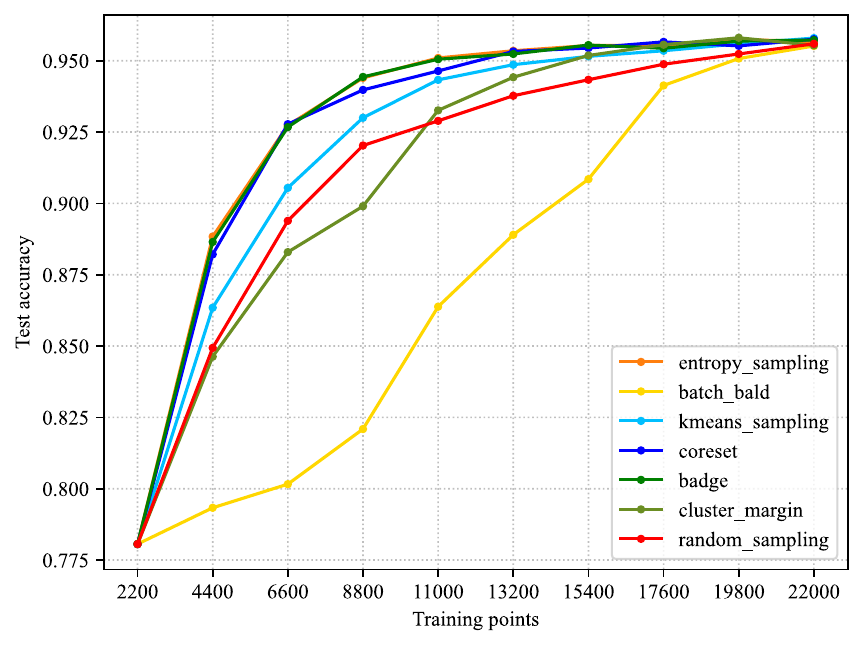}}
	\subfigure[EuroSAT Self-SL]{\includegraphics[width=0.245\linewidth]{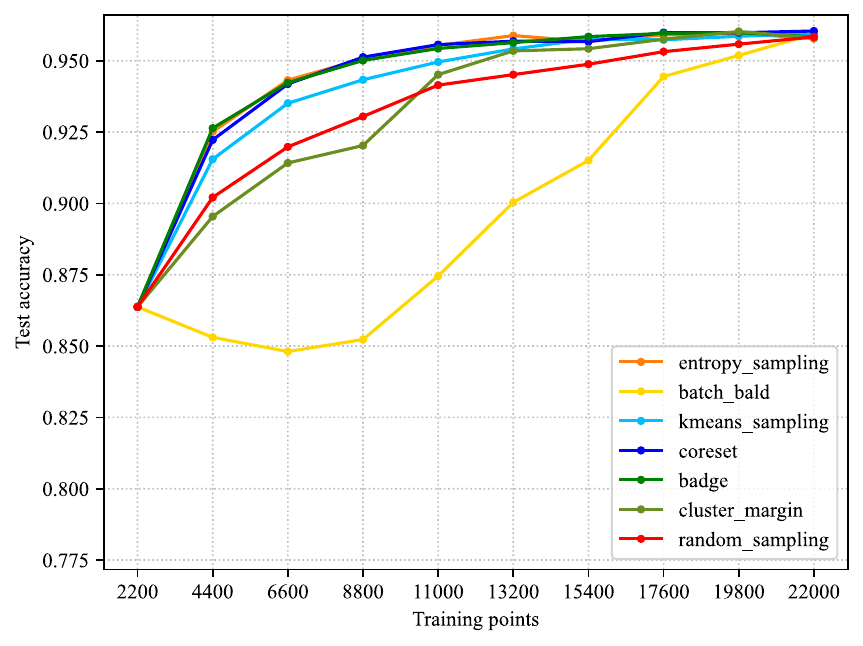}}
     \subfigure[OCT scratch]
     {\includegraphics[width=0.245\linewidth]{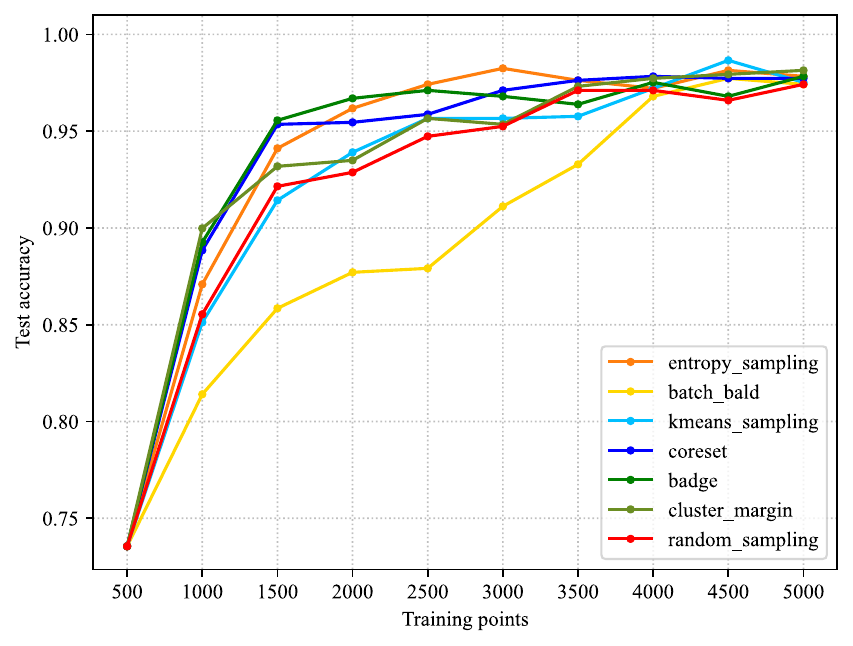}}
	\subfigure[OCT Self-SL]{\includegraphics[width=0.245\linewidth]{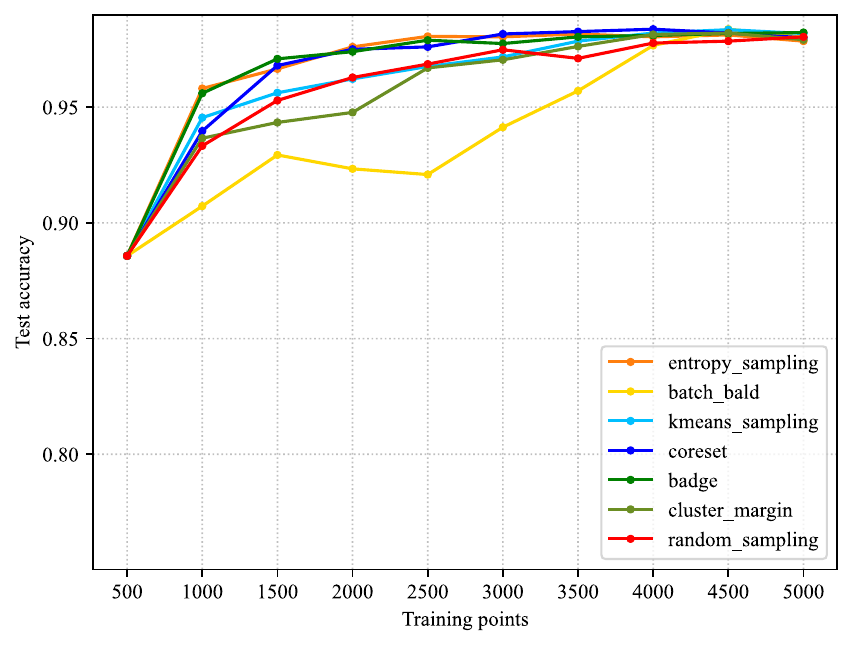}}
     \subfigure[BrainTumor scratch]{\includegraphics[width=0.245\linewidth]{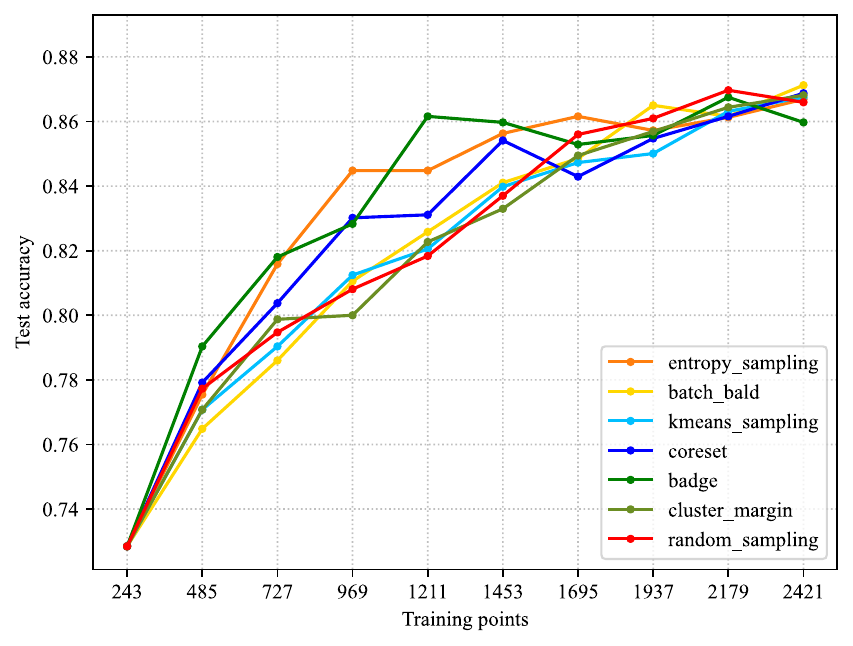}}
	\subfigure[BrainTumor Self-SL]{\includegraphics[width=0.245\linewidth]{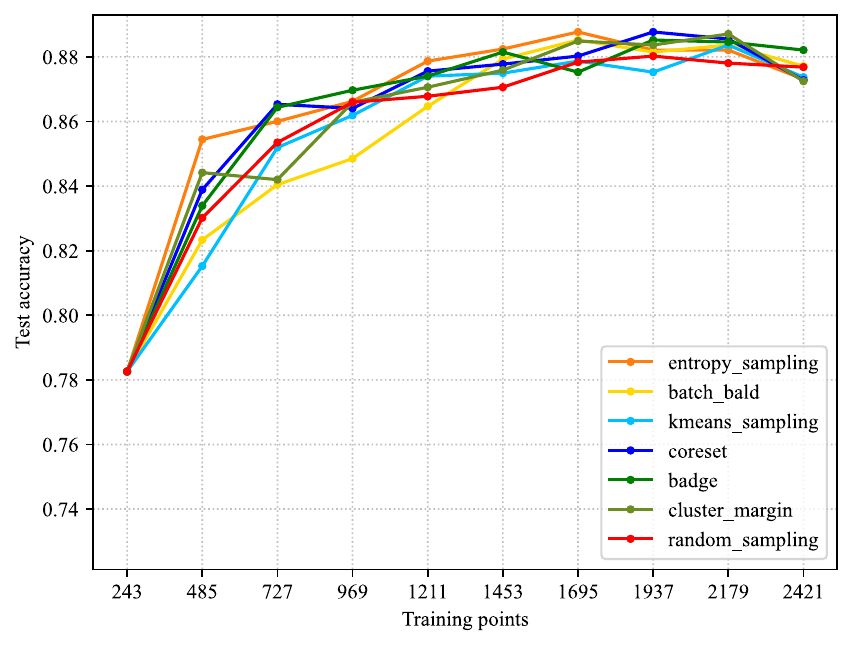}}
     \subfigure[GAPs scratch]
     {\includegraphics[width=0.245\linewidth]{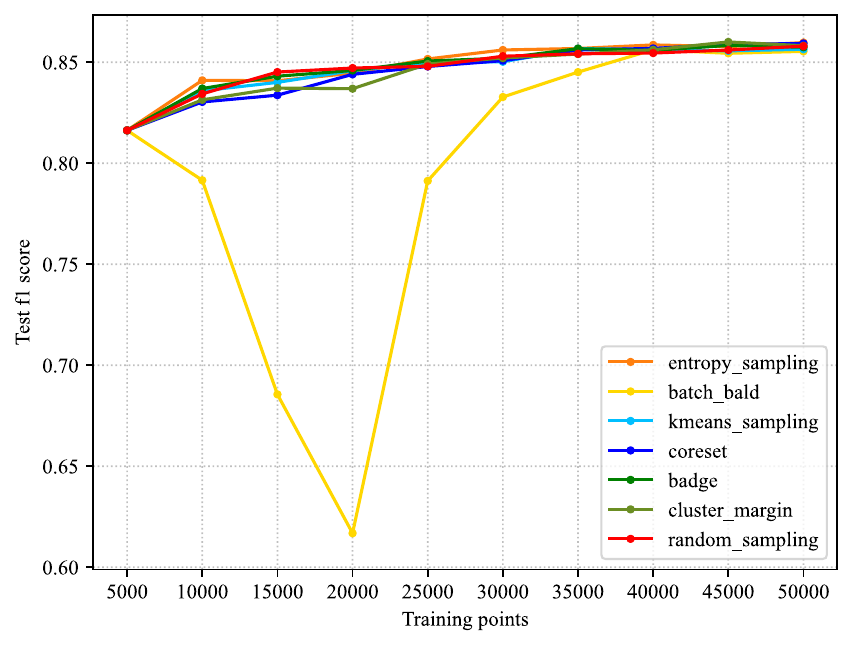}}
	\subfigure[GAPs Self-SL]{\includegraphics[width=0.245\linewidth]{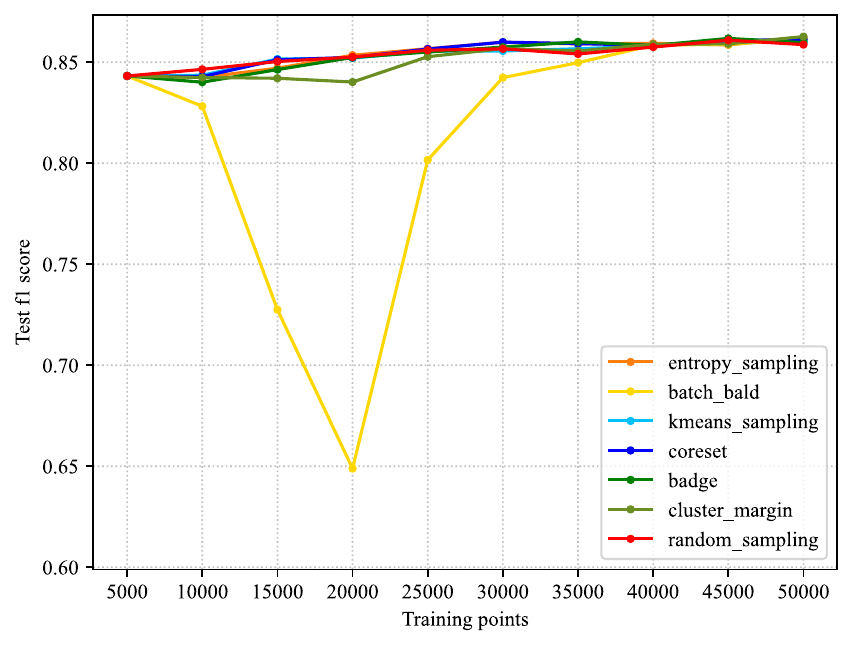}}
     \subfigure[KolektorSDD2 scratch]{\includegraphics[width=0.245\linewidth]{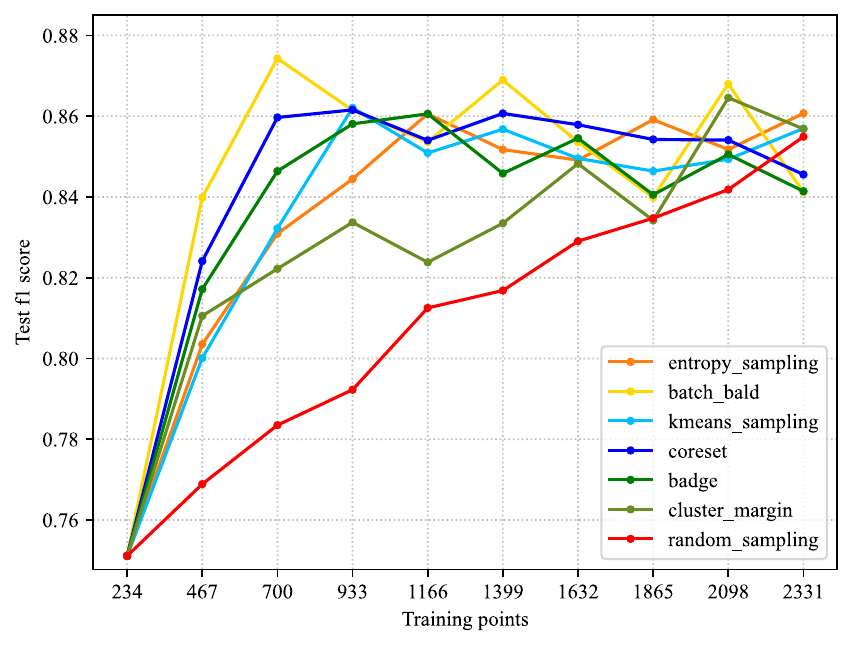}}
	\subfigure[KolektorSDD2 Self-SL]{\includegraphics[width=0.245\linewidth]{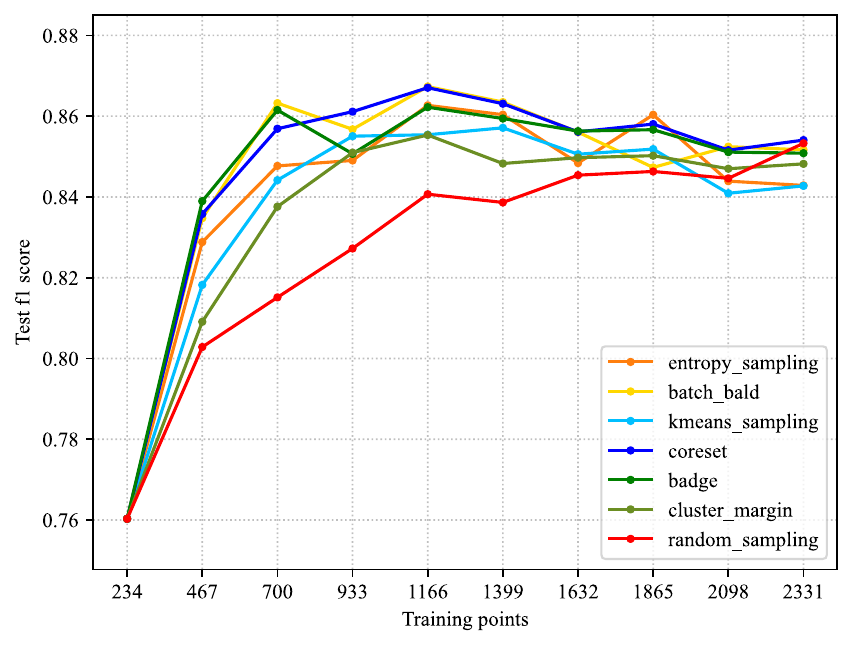}}
    \caption{Result of benchmark. The horizontal axis represents the number of annotated samples and the vertical axis represents the corresponding test accuracy(or F1-score). By using query strategies, the model can reach the final accuracy of Random Sampling earlier, indicating that annotation cost can be reduced.}
\label{result}
\end{figure*}

The results of the experiments are shown in Figure~\ref{result}. 
The horizontal axis represents the number of annotated samples and the vertical axis represents the corresponding test accuracy~(or F1-score) . 
By using query strategies, the model can reach the final accuracy of Random Sampling earlier, indicating that annotation cost can be reduced.

For CIFAR10, Entropy Sampling, Core-set, and BADGE reach the same accuracy at 35,000 data as that obtained training on all data, resulting in a 30\% reduction in annotation cost.
K-means Sampling and Cluster Margin reach the same accuracy at 40,000 data as that obtained training on all data, resulting in a 20\% reduction in annotation cost.
In contrast, BatchBALD performs similarly to Random Sampling and thus does not reduce the annotation cost.
These results indicate that for CIFAR10, Entropy Sampling, Core-set, and BADGE are the most effective methods, followed by K-means Sampling and Cluster Margin, while BatchBALD performs poorly. The same results were observed for EuroSAT.

For OCT and BrainTumor, Entropy Sampling, Core-set, and BADGE showed high performance. The other query strategies had the same performance as that of Random Sampling. This suggests that some query strategies are effective even for medical images.

For GAPs, the score did not improve even as the number of labeled data $|\mathcal{L}_c|$ increased.
This is because GAPs contains noisy data, and DAL cannot consider noisy labels during the query phase because annotation is performed after querying. Therefore, robust learning methods such as datacleansing~\cite{ReviewDataCleansingMethodsBigData,Learningnoisylabelsdeepneuralnetworkssurvey} or framework improvements are required.

\begin{figure}[!tb]
\centering
    \subfigure[(a) BatchBALD]{\includegraphics[width=0.45\linewidth]
    {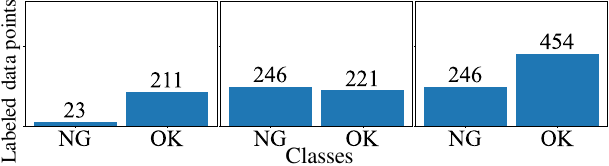}}
     \subfigure[(b) Entropy Sampling]{\includegraphics[width=0.45\linewidth]{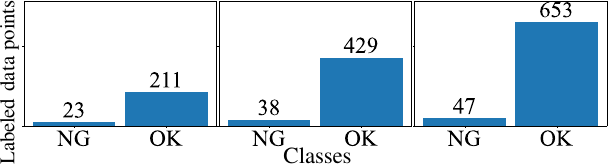}}
    \caption{Distribution of training data in the first three cycles of KolektorSDD2. 
    With BatchBALD, a large number of NG samples were queried early on, resulting in balanced learning and a significant improvement in performance in the early stages. In contrast, with Entropy Sampling, a large number of OK samples were queried, resulting in more imbalanced learning and a decrease in performance.}
\label{kolektor_hist}
\end{figure}

For KolektorSDD2, BatchBALD obtained the highest score at 700 labeled samples. 
To investigate the reason for this, we show the distribution of labeled data (from 234 to 700 samples) for the high-performing BatchBALD and the low-performing Entropy Sampling in Figure~\ref{kolektor_hist}. 
As shown, BatchBALD queries NG samples early on, which mitigates the data imbalance, whereas Entropy Sampling queries a large number of OK samples, resulting in an imbalance that is not alleviated and an initial lack of accuracy improvement in the cycle. 
However, since BatchBALD performs poorly on other datasets, it can only be used in limited problem settings, such as imbalanced binary classification~(e.g., KolektorSDD2).

Except for visual inspection images, self-SL was found to be effective in reducing annotation cost compared to scratch, indicating the effectiveness of self-SL in DAL. 
Self-SL did not work well for visual inspection images, likely due to the small variation in the anomaly-free data of inspection images, which resulted in pre-training failure.
For SimSiam, different data augmentations are applied to each sample to acquire two augmented data and learning is performed to increase their similarity. However, since anomaly-free data are similar to each other, the augmented data obtained from these data are also similar, and it may thus not be possible to sufficiently learn representations.
Therefore, further performance improvement for visual inspection images is expected if pre-training using approaches other than contrastive learning is applied.

For CIFAR10 and EuroSAT, the performance of Entropy Sampling, Core-set, and BADGE are very close. 
This suggests that there is no difference in performance between these query strategies for these datasets, indicating the limits of existing DAL methods.
The development of methods and the verification of their effectiveness for these datasets are thus difficult. In the next section, we conduct verification experiments.

\section{Verification Experiments: Querying Using Fully-trained Model}
In this section, we conduct verification experiments to investigate the effectiveness of the current DAL approach for various datasets.
The performance limit of DAL is obtained by learning with the optimal labeled data $\mathcal{L}_c^*$ in each cycle. 
However, obtaining the optimal $\mathcal{L}_c^*$ requires solving a combinatorial problem, which is impractical. 
Therefore, we focus on the model-dependent approach of querying based on the model's prediction probabilities or features in the current DAL method and modify the framework.

\subsection{Framework and Settings for Verification Experiments}
\begin{figure}[!tb]
    \begin{center}
    \includegraphics[keepaspectratio, width=70mm]{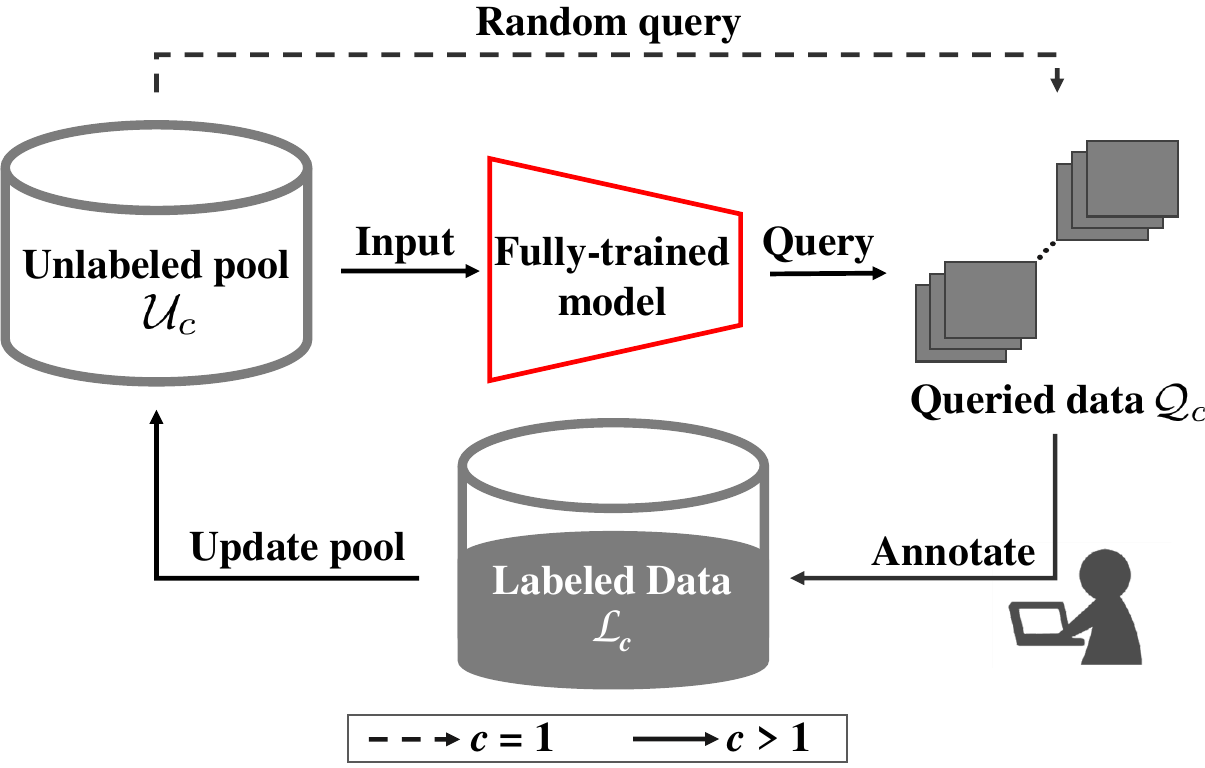}
    \caption{Framework for verification experiments}
    \label{cheater}
    \end{center}
\end{figure}

The framework used for the verification experiments is shown in Figure~\ref{cheater}. 
In conventional DAL, querying is done by a model trained on $\mathcal{L}_c$. However, when $|\mathcal{L}_c|$ is small, the model's learning can be insufficient, resulting in poor feature representations.
Query strategies using such feature representations may not work sufficiently, and it may not be possible to query high-quality $\mathcal{Q}_c$.
Therefore, we replace the model used for querying with a fully-trained model~(i.e., a model trained on the entire training data) to query $\mathcal{Q}_c$ that is independent of $|\mathcal{L}_c|$.
We investigate whether current model-dependent DAL methods improve their performance by comparing the results of the verification experiments with the benchmark results presented in Section~\ref{result of benchmark} and Figure~\ref{cheater}.

\begin{table}[!tb]
\caption {\textmd{Performance of fully-trained model for various datasets}}
\vspace{3.0mm}
\label {cheat_models}
\centering
\scalebox{0.8}{
    \begin{tabular}{lcc}
    \hline
    Dataset         & Evaluation metric         & Performance      \\ \hline
    CIFAR10         & Test accuracy             & 95.48\%          \\
    EuroSAT         & Test accuracy             & 95.18\%          \\
    OCT             & Test accuracy             & 98.24\%          \\
    BrainTumor      & Test accuracy             & 89.26\%          \\
    GAPs            & Test f1 score             & 85.33\%          \\
    KolektorSDD2    & Test f1 score             & 89.80\%          \\ \hline
    \end{tabular}
}
\end{table}

In the verification experiments, we use the same datasets as those in the benchmark and employ the query strategies that performed well: Entropy Sampling, Core-set, and BADGE. 
In addition, we use BatchBALD for KolektorSDD2 because it showed the best performance. We use the same hyperparameters as those in the benchmark and only perform learning from scratch. The performance of the fully-trained models used in each dataset is shown in Table~\ref{cheat_models}.

\begin{figure*}[!tb]
\centering
    \subfigure[CIFAR10]{\includegraphics[width=0.3\linewidth]
    {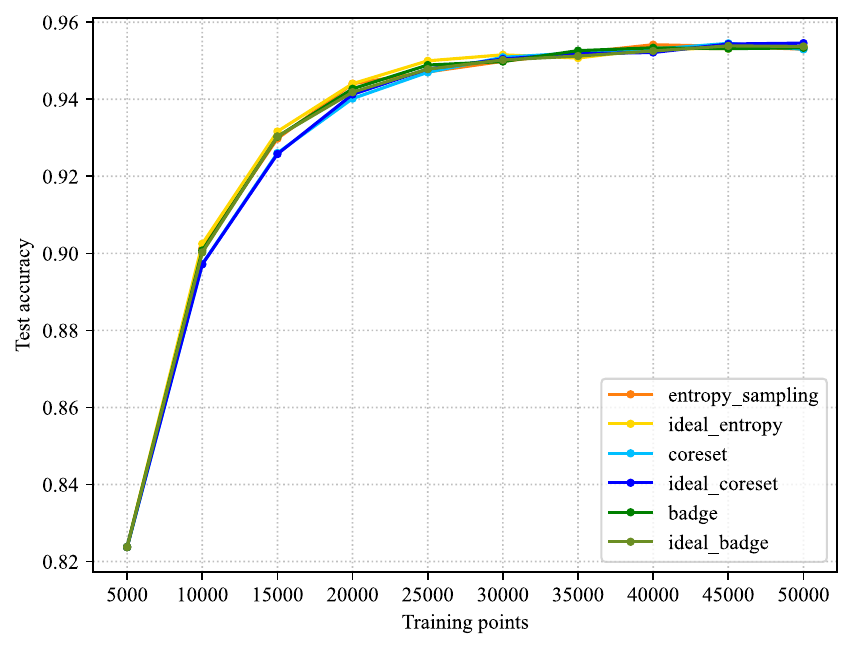}}
     \subfigure[EuroSAT]{\includegraphics[width=0.3\linewidth]{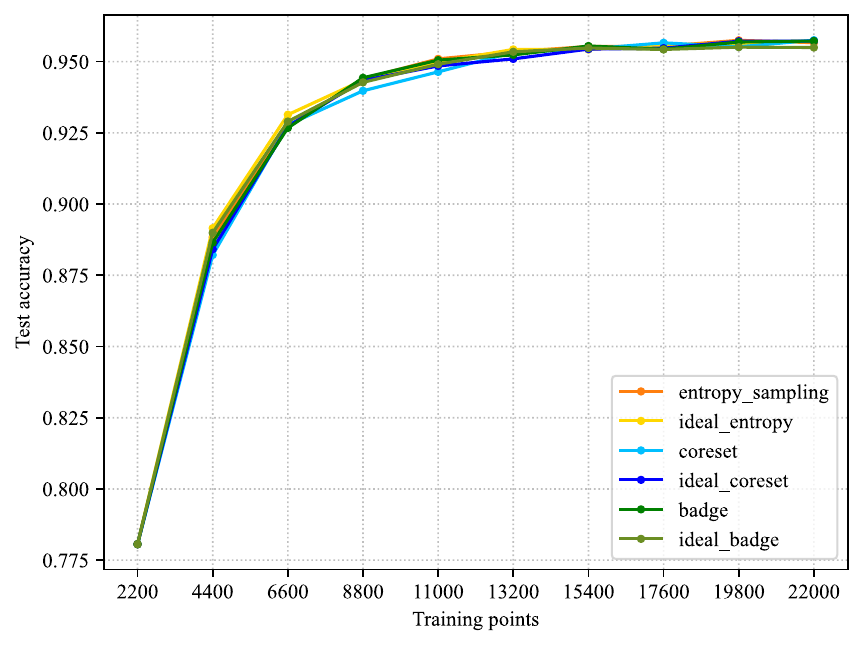}}
     \subfigure[OCT]
     {\includegraphics[width=0.3\linewidth]{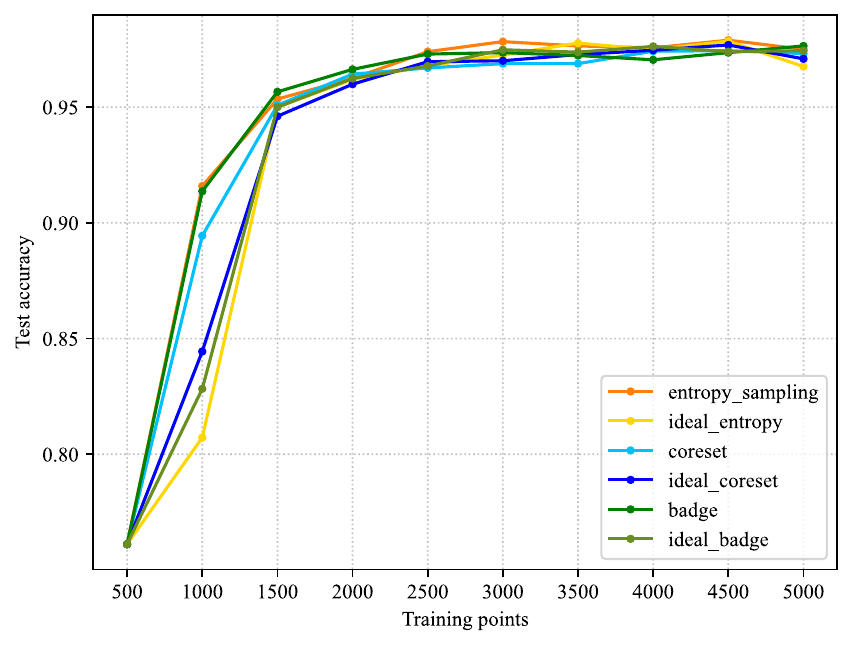}}
     \subfigure[BrainTumor]{\includegraphics[width=0.3\linewidth]{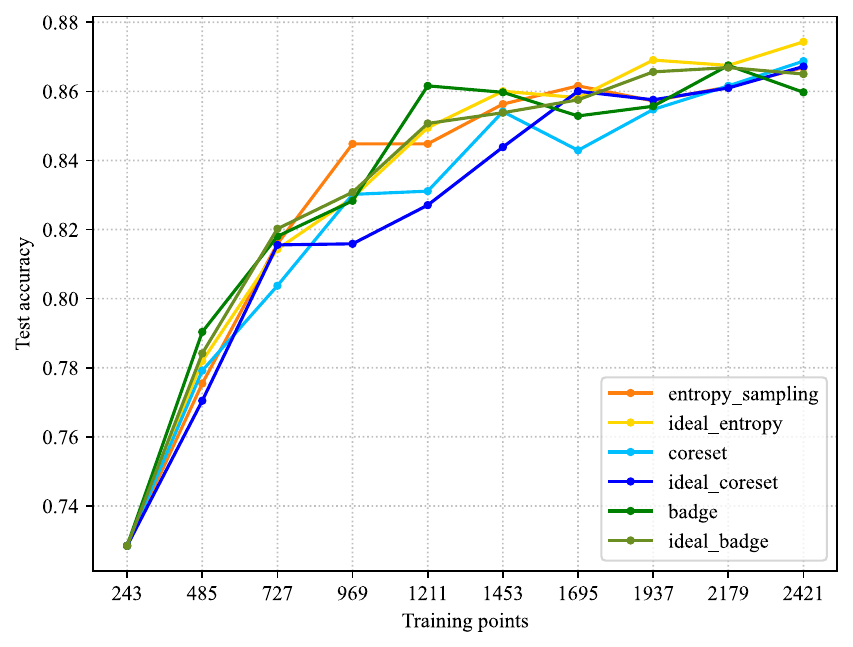}}
     \subfigure[GAPs]
     {\includegraphics[width=0.3\linewidth]{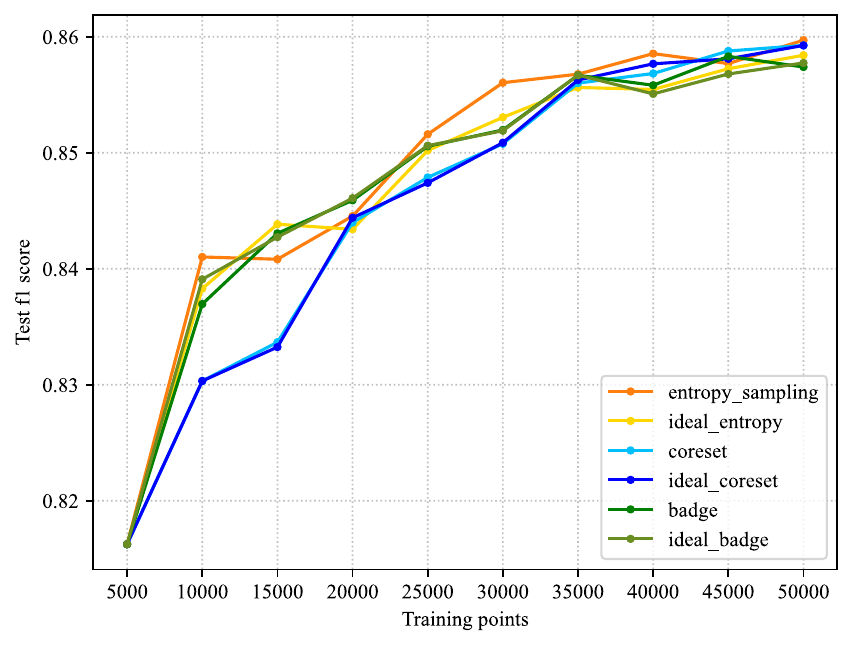}}
     \subfigure[KolektorSDD2]{\includegraphics[width=0.3\linewidth]{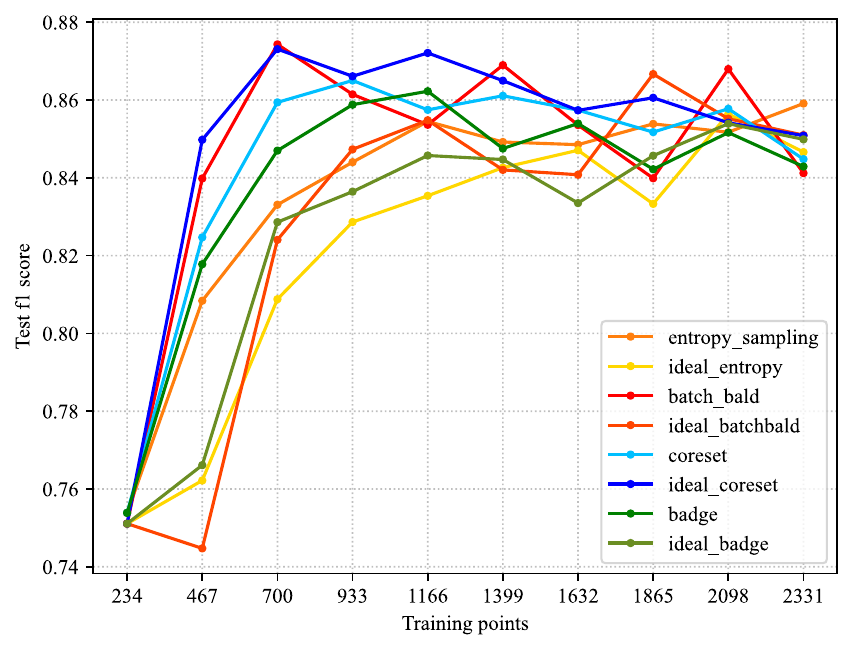}}
    \caption{Results of verification experiments. Results of ideal\_entropy, ideal\_coreset, ideal\_badge, and ideal\_batchbald under the framework shown in Figure~\ref{cheater}.}
    
\label{cheating}
\end{figure*}

\subsection{Results of Verification Experiments}
The results of the verification experiments are shown in Figure~\ref{cheating}. As shown, there is no performance difference between the regular framework and the framework shown in Figure~\ref{cheater} for CIFAR10 and EuroSAT. However, performance differences were observed for other datasets, such as KolektorSDD2.

DAL achieves high performance with limited $|\mathcal{L}_c|$ by constructing $\mathcal{L}_c$ with only high-quality data. 
Therefore, no performance difference was observed for homogeneous datasets such as CIFAR.
On the other hand, for non-homogeneous datasets such as KolektorSDD2, performance differences were observed.

Thus, further development of DAL is expected if query strategies for datasets such as KolektorSDD2 are proposed. For designing such query strategies, it is necessary to consider the characteristics and purpose of the given dataset.
\section{Conclusions}
In this paper, we showed the problems of existing studies on DAL, such as non-standardized hyperparameters and limited datasets, and demonstrated that due to these problems, it is difficult to compare query strategies and evaluate the performance of DAL on medical or visual inspection images.
We then developed standardized experimental settings as a baseline for future DAL research and investigated the effectiveness of various query strategies.

In benchmark experiments conducted to examine the effectiveness of existing query strategies, we confirmed their effectiveness not only on natural images but also on satellite, medical, and visual inspection images. Poor performance was observed for datasets that include noisy labels, such as GAPs, for which we confirmed that DAL itself does not function properly. 
Since annotation errors may occur in practical use, it is necessary to propose robust frameworks or learning methods for noisy labels.
Furthermore, for all datasets except for visual inspection images, we confirmed that the use of self-SL in pre-training improves accuracy and increases the annotation cost reduction that can be achieved by DAL.
However, we could not confirm the effectiveness of Self-SL on visual inspection images. This is thought to be due to the high similarity between anomaly-free data in visual inspection images. Therefore, applying effective self-SL methods to visual inspection images is expected to further reduce annotation costs.

Next, to investigate the effectiveness of the current DAL, we conducted verification experiments that compared the results obtained using a framework that uses a fully-trained model at each cycle during query and those obtained using a normal framework for existing datasets to verify for which datasets DAL is effective.
The results of the verification experiments confirm that there is no performance difference for homogeneous datasets such as CIFAR10 and EuroSAT among the query strategies. In contrast, there was a performance difference for non-homogeneous datasets such as KolektorSDD2. 
This indicates that many current DAL methods do not show a significant improvement in performance for homogeneous datasets. These methods are not expected to improve significantly in the future.
However, there is great potential for new query strategies for non-homogeneous datasets, which are encountered in practical use. 
Therefore, in the future, we need to develop query strategies that consider the characteristics and purpose of the given dataset to increase the effectiveness of DAL in practical applications.

\bibliographystyle{unsrtnat}
\bibliography{main}
\end{document}